# A Matlab Toolbox for Feature Importance Ranking*

Shaode Yu, Zhicheng Zhang, Xiaokun Liang, Junjie Wu, Erlei Zhang, Wenjian Qin, and Yaoqin Xie

*Abstract* - More attention is being paid for feature importance ranking (FIR), in particular when thousands of features can be extracted for intelligent diagnosis and personalized medicine. A large number of FIR approaches have been proposed, while few are integrated for comparison and real-life applications. In this study, a matlab toolbox is presented and a total of 30 algorithms are collected. Moreover, the toolbox is evaluated on a database of 163 ultrasound images. To each breast mass lesion, 15 features are extracted. To figure out the optimal subset of features for classification, all combinations of features are tested and linear support vector machine is used for the malignancy prediction of lesions annotated in ultrasound images. At last, the effectiveness of FIR is analyzed according to performance comparison. The toolbox is online (https://github.com/NicoYuCN/matFIR). In our future work, more FIR methods, feature selection methods and machine learning classifiers will be integrated.

## I. INTRODUCTION

More attention is being paid for feature importance ranking (FIR), in particular when thousands of features can be crafted for intelligent diagnosis and personalized medicine [1]. In the field of radiomics, high-throughput features can be collected from medical images regarding the analysis of intensity statistics, shape description, texture analysis and distributions of gradient, and such computational features in multi-scale spaces, multiple transform domains, and multi-modality image acquisitions [2, 3, 4].

Thousands of features can be computed and how to retrieve informative and discriminative ones is still on pending [5]. FIR plays an important role for feature selection and follow-up information processing. The purpose of FIR is to sort features according to the relative importance in a specific measure space. In general, FIR methods can be grouped into supervised and unsupervised methods depending on whether the labels of training samples are used, while to the supervised methods, labels are employed [6]. From the technical perspective, some methods use statistical analysis and class separability criteria, such as two-sample *t*-test and Wilcoxon rank-sum test, to quantify inter-feature relationship; some methods explore the mutual information [7], joint embedding learning and spare regression, structured graph optimization and nonnegative spectral analysis; while some take classification performance into account and the selection of a machine learning classifier becomes important [6].

A large number of FIR methods have been proposed, while few are integrated for comparison and real-world applications. To our knowledge, only one toolbox, Feature Selection Library (FSLib), collects 19 algorithms implemented on MATLAB [8]. In this study, a matlab toolbox is presented and a total of 30 FIR methods are integrated. Further, the toolbox is tested on a database of 163 ultrasound images [9]. To each breast mass lesion, 15 features are extracted. To figure out the optimal subset of features, all combinations of features are evaluated and linear support vector machine (SVM) is used for the prediction of lesion malignancy. At last, the effectiveness of FIR is analyzed according to performance comparison. The toolbox is online (https://github.com/NicoYuCN/matFIR).

The remainder of this paper is structured as follows. Section II provides a brief review of the 30 FIR methods, data collection, feature extraction and experiment design. Section III shows and discusses the ranking results and at last, Section IV concludes this study.

## II. MATERIALS AND METHODS

### A. A brief review of FIR methods

A total of 30 methods have been integrated. (1) *fir_asl*. It provides a unified learning framework and performs structure learning and feature selection simultaneously. Iteratively, it is able to capture accurate structures and select more informative features [10]. (2) *fir_cor*. It assumes that good feature subsets contain features are highly correlated with the classification labels, while the features are uncorrelated to each other [11]. (3) *fir_dgufs*. It is a projection-free model and aims to enhance the inter-dependence among original data, clustering labels and selected features [12]. (4) *fir_ec*. It is graph-based and ranks features by identifying the most important ones into arbitrary set of cues (i.e. eigenvector centrality) [13]. (5) *fir_fisher*. It finds a subset of features by maximizing the lower bound of Fisher score and the problem is formulated as a quadratically constrained linear programming [14]. (6) *fir_fsv*. It considers all the possible subsets of features and improves the robustness of the selected features and is a probabilistic latent graph-based algorithm [15]. (7) *fir_gini*. It tends to balance concentration and dependency ratios for feature selection [16]. (8) *fir_glsi*. It provides a unified framework based on spectral graph theory and it can exploit the intrinsic properties underlying supervised

* This work is supported in part by grants from the Shenzhen matching project (GJHS20170314155751703), the National Key Research and Develop Program of China (2016YFC0105102), the National Natural Science Foundation of China (61871374), the Leading Talent of Special Support Project in Guangdong (2016TX03R139), the Fundamental Research Program of Shenzhen (JCYJ20170413162458312), the Science Foundation of Guangdong (2017B020229002, 2015B020233011, 2014A030312006) and CAS Key Laboratory of Human-Machine Intelligence-Synergy Systems, Shenzhen Engineering Laboratory for Key Technologies on Intervention Diagnosis and Treatment Integration

S Yu was with the Shenzhen Institutes of Advanced Technology, Chinese Academy of Sciences, Shenzhen, GD 518055, China. He is now with the Radiation Oncology Department, University of Texas Southwestern Medical Center, Dallas, TX 75390 USA (e-mail: shaode.yu@utsouthwestern.edu).

X Liang, Z Zhang, W Qin and Y Xie are with the Shenzhen Institutes of Advanced Technology, Chinese Academy of Sciences, Shenzhen, GD 518055, China (corresponding author, Y Xie; phone: 086-755-8639-2281; fax: 086-755-8639-2299; e-mail: yq.xie@siat.ac.cn).

J Wu is with the Lyle School of Engineering, Southern Methodist University, Dallas, TX 75205, USA (email: junjie@smu.edu).

E Zhang is with the College of Information Science and Technology, Northwest University, SX 710069, China (email: erlei.zhang@nwu.edu.cn).

and unsupervised fashion [17]. (9) *fir_il*. It models relevancy as a latent variable in a probabilistic latent semantic analysis inspired generative process [18]. (10) *fir_inf*. It concerns a selection of features as a path among feature distributions and letting these paths tend to an infinite number. The importance (relevance and redundancy) of a feature is investigated when injected into an arbitrary set of cues [19]. (11) *fir_jelsr*. It is an unsupervised feature selection approach via joint embedding learning and sparse regression. Moreover, it weights features via locally linear approximation to construct graph and unify embedding learning and sparse regression [20]. (12) *fir_KW*. The optimal feature subset is learnt from data independent of parameter settings [21]. (13) *fir_lapscore*. It highlights itself by evaluating the importance of a feature from its power of locality preserving (Laplacian score) [22]. (14) *fir_llc*. It uses kernel learning within the framework of local learning-based clustering to obtain an appropriate data representation [23]. (15) *fir_lnrd*. It assumes that the class labels of input data can be predicted by a linear classifier and both discriminative analysis and the $L_{2,1}$-norm minimization are incorporated into a joint framework for feature selection [24]. (16) *fir_mat_ttest*. It sorts features by using class separability criteria of two-sample *t*-test [25]. (17) *fir_mat_entropy*. The class separability criteria of relative entropy, also known as Kullback-Leibler distance or divergence, is used for feature ranking [26]. (18) *fir_mat_bhy*. Features are ranked with regard to class separability criteria of minimum attainable classification error which is also known as the Chernoff bound [27]. (19) *fir_mat_roc*. It ranks feature based on the class separability criteria of the area between the empirical receiver operating characteristic curve (EROC) and the random classifier slope [28]. (20) *fir_mat_wilcoxon*. It sorts features based on the class separability criteria of absolute value of the standardized *u*-statistic of a two-sample unpaired Wilcoxon test or Mann-Whitney [29]. (21) *fir_mat_relieff*. It is an update of Relief and it finds that absolute differences are sufficient to update the weight vector rather than using the square of value differences [30]. (22) *fir_mat_lasso*. It aims to improve the accuracy and interpretability of regression models by altering the model fitting process to select only a subset of the provided covariates for use in the final model [31]. (23) *fir_mc*. It uses manifold learning and $L_1$-regularized models to select a subset of features and meanwhile the multi-cluster structure of the data is preserved [32]. (24) *fir_mmls*. It tries to minimize the within-locality information of the same classes and meanwhile to maximize the between-locality information of the different classes [33]. (25) *fir_nnsa*. It performs spectral clustering to learn the cluster labels of input samples and meanwhile, features are selected [34]. (26) *fir_ol*. It designs a triplet-based loss function for feature selection that preserve ordinal locality of original data and besides, the orthogonal basis clustering is simplified by imposing a constraint on the feature projection matrix [35]. (27) *fir_pwfp*. It provides a pair-wise feature selection for high dimension but low sample size data [36]. (28) *fir_ru*. It utilizes robust nonnegative matrix factorization, local learning and robust feature learning into feature selection [37]. (29) *fir_sgo*. It simultaneously performs local structure learning and feature selection, and the similarity matrix can be determined adaptively [38]. (30) *fir_soc*. Latent cluster information is used by the target matrix conducting orthogonal basis clustering in a single unified term of the proposed objective function which can be minimized by a simple optimization algorithm [39].

### B. Data collection

An ultrasound breast image dataset (UDIAT) is analyzed in this study [9]. It contains 163 B-mode images using 8.5 MHz transducer. The average size of image matrix is [760 570] and the physical resolution is 0.084 mm. Breast mass lesions were annotated by experienced radiologist. Histological verification indicates that 54 mass lesions are malignant and 109 lesions are benign.

### C. Feature extraction

To each mass lesion, 15 features are collected. Specifically, 7 features quantify mass intensity (mean, median, standard deviation, maximum, minimum, kurtosis and skewness), 5 features represent the lesion shape, including area, perimeter, circularity, elongation and form, and 3 features describe the texture (contrast, correlation and entropy) based on grey-level co-occurrence matrix (GLCM) [40].

Features computed from intensity statistics. Let $X$ denotes the pixel intensities in a lesion with $n$ pixels ($X = \{x_i\}_{i=1}^{n}$) in it and $x_i$ is the intensity value of the $i^{th}$ pixel. The mean value ($\mu$) is defined as $\mu = \frac{1}{n}\sum_{i=1}^{n} x_i$ and the bias ($\beta_i$) of $x_i$ to the $\mu$ defined as $\beta_i = x_i - \mu$.

(1) $i_{\text{mean}} = \mu$

(2) $i_{\text{median}}$

(3) $i_{\text{std\_dev}} = (\frac{1}{n-1}\sum_{i=1}^{n}(\beta_i)^2)^{1/2}$

(4) $i_{\text{maximum}}$

(5) $i_{\text{minimum}}$

(6) $i_{\text{kurtosis}} = \dfrac{\frac{1}{n}\sum_{i=1}^{n}(\beta_i)^4}{(\sqrt{\frac{1}{n}\sum_{i=1}^{n}(\beta_i)^2})^2}$

(7) $i_{\text{skewness}} = \dfrac{\frac{1}{n}\sum_{i=1}^{n}(\beta_i)^3}{(\sqrt{\frac{1}{n}\sum_{i=1}^{n}(\beta_i)^2})^3}$

Features computed from shape description. To each lesion region, an ellipse is fitted based on the linear least square. Here, $m$ represents the length of the minor axis, while $M$ denotes the length of the major axis of the ellipse. In addition, the ellipse has the same normalized second central moments as the region surrounded by the contour.

(8) $s_{\text{area}}$

(9) $s_{\text{perimeter}}$

$$(10)\ s_{circularity} = 4\pi \frac{s_a}{(s_p)^2}$$

$$(11)\ s_{elongation} = \frac{m}{M}$$

$$(12)\ s_{form} = \frac{s_p \times s_e}{8 \times s_a}$$

Features computed from texture analysis. For quantification of texture analysis, GLCM is conducted $P(i, j; L, d, \alpha)$. In this study, the gray-level quantization ($L$), the direction ($\alpha$) and the distance ($d$) as to the matrix of GLCM is defined as 32, $0°$ and 1 pixel, respectively. Then, the GLCM $P(i,j)$ is normalized by the sum of its elements to calculate the co-occurrence relative frequency between the gray level of $i$ and $j$. Besides, $p_x(i) = \sum_{j=1}^{L} P(i,j)$ is defined as marginal row probabilities, $p_y(j) = \sum_{i=1}^{L} P(i,j)$ as marginal column probabilities, $\mu_x$ as the mean of $p_x$, $\mu_y$ as the mean of $p_y$, $\sigma_x$ as the mean of $p_x$ and $\sigma_y$ as the mean of $p_y$.

$$(13)\ t_{contrast} = \sum_{i=1}^{L} \sum_{j=1}^{L} |i-j|^2 P(i,j)$$

$$(14)\ t_{correlation} = \frac{\sum_{i=1}^{L} \sum_{j=1}^{L} i \times j \times P(i,j) - \mu_x \mu_y}{\sigma_x \sigma_y}$$

$$(15)\ t_{entropy} = -\sum_{i=1}^{L} \sum_{j=1}^{L} P(i,j) \log_2(P(i,j))$$

*D. Experiment design*

It is challenging on evaluating the effectiveness of a FIR method in a quantitative way. This study uses a straightforward strategy. It considers all combinations of feature subsets (the maximum number of features is 8). Based on an exhaustive search, linear support vector machine (SVM) [41] performs as the classifier for malignancy prediction of lesions in ultrasound images, since linear models are easy for interpretation.

Figure 2 shows the linear SVM which attempts to find a hyper-plane (known as the maximum-margin hyperplane) that can represent the largest separation, or margin, between the two classes. In an exhaustive evaluation, the feature subset that achieves the maximum value of the area under the curve (AUC) is defined as the optimal subset.

Therefore, the effectiveness of a method for feature ranking is defined as $eff = \frac{m}{n}$ where $m$ is the number of features in the optimal subset, while $n$ is the minimum number of ranked features that contains the selected $m$ features.

## III. RESULTS AND DISCUSSION

Ranking features according to their relative importance in data classification and numerical regression is an overlooked topic, in particular when deep learning dominates the field of artificial intelligence [42, 43, 44]. In the era of big data, information overloaded imposes difficulties on understanding an issue and subsequently, it is challenging to make decisions effectively. Thus, FIR becomes more important and it attempts to figure out the informative, representative and discriminative features that can describe intrinsic characteristics of the issue. This study presents a matlab toolbox that integrates up to 30 FIR methods. Furthermore, an evaluation on ultrasound image based breast lesion classification is conducted. To quantify the effectiveness, an index is proposed.

Based on the database UDIAT, an exhaustive search of all possible feature subsets (note that the maximum number is 8) is evaluated and linear SVM is the classifier for breast lesion classification. It is found that 3 features can achieve the best performance and the AUC is $0.756 \pm 0.037$. These features are $i_{median}$ (index 2), $i_{skewness}$ (index 7) and $t_{contrast}$ (index 13). The former 2 features are computed from intensity statistics and the last feature describes image contrast. It might indicate that to breast lesion diagnosis when using ultrasound images, visual information is more crucial, such as image intensity and tissue contrast. Interestingly, these features are perceivable and thus, decision making becomes more direct. On contrast, features from shape description is not included in the finding. It is known that shapes are dependent on the annotation, while the annotation is highly dependent on the radiologists' knowledge and experience. And thus, shape features might be not reliable nor reproducible. Moreover, the linear SVM based best result is quite close to deep learning based result [45].

Table 1 shows the feature ranking results which shows the FIR methods, ranked indexes of features and the effectiveness. Note that the red index in bold is the index of the third feature in the optimal subset. It is observed different FIR methods give out different ranking results due to different ranking principles. At first, no method ranks the top 3 features as the exhaustive search method finds. The best ranking method comes from the method *fir_mat_ttest* (3/4) and *fir_fisher* (3/4), followed by *fir_ec* (3/5), *fir_inf* (3/5) and *fir_glsi* (3/6), and other methods require up to 7 features to include the optimal subset. It is also observed that 5 out of 30 FIR methods (*fir_asl*, *fir_gini*, *fir_mat_lasso*, *fir_nnsa* and *fir_ol*) lead to poor ranking results. It indicates that some FIR methods are not able to figure out the most important features in terms of this UDIAT database.

This study has some limitations. First, for further enriching our understanding of FIR methods, it is better to classify these methods into different groups based on the principle of how to weight feature importance. For example, some are based on class separability criteria [25, 26, 27, 28, 29], others weight feature importance in graph space [13, 17, 20], while some aims to preserve the local data structure [20, 22, 23, 33, 35, 37, 38]. Second, one database is involved in this study and more datasets with regard to different purpose should be evaluated. As such, the effectiveness might be fairly compared. Note that the numbers of benign and malignant lesions are not balanced that might lead to biased ranking according to features' relative importance [46]. Third, more classifiers can be further tested, such as artificial neural network, naïve Bayes, decision tree, random forest, extreme learning machine (ELM), nonlinear SVM and ensemble approaches [47, 48, 49]. Besides, high through features can be collected in this study and the number of features can be represented in multi-scale space, or multiple transform domains [1, 2, 3, 4]. That definitely increases time cost and it also makes this study more complicated and hard to compare. Last but not the least, it is challenging to measure the effectiveness of a FIR method which is not only related to time cost but also the specific purpose. And thus, attention should be paid in our future work for the effectiveness analysis of FIR.

TABLE I. FIR RESULTS ON THE UDIAT DATABASE FOR ULTRASOULD LESION CLASSIFICATION

| FIR methods | Index for ranked features | | | | | | | | | | | | | | | eff |
|---|---|---|---|---|---|---|---|---|---|---|---|---|---|---|---|---|
| fir_asl | 6 | 1 | 2 | 3 | 14 | 5 | 4 | 11 | 9 | 7 | 8 | 10 | 15 | 13 | 12 | 3/14 |
| fir_cor | 4 | 6 | 8 | 7 | 5 | 3 | 15 | 12 | 9 | 11 | 2 | 13 | 1 | 10 | 14 | 3/12 |
| fir_dgufs | 13 | 12 | 1 | 2 | 3 | 4 | 5 | 6 | 7 | 8 | 9 | 10 | 11 | 14 | 15 | 3/9 |
| fir_ec | 8 | 7 | 6 | 13 | 2 | 1 | 15 | 11 | 5 | 14 | 12 | 9 | 3 | 4 | 10 | 3/5 |
| fir_fisher | 6 | 7 | 13 | 2 | 11 | 1 | 15 | 14 | 12 | 8 | 3 | 9 | 10 | 4 | 5 | 3/4 |
| fir_fsv | 14 | 12 | 13 | 10 | 15 | 7 | 6 | 11 | 2 | 1 | 5 | 3 | 4 | 9 | 8 | 3/8 |
| fir_gini | 5 | 14 | 10 | 4 | 8 | 3 | 9 | 15 | 11 | 1 | 2 | 12 | 13 | 7 | 6 | 3/14 |
| fir_glsi | 5 | 12 | 6 | 13 | 7 | 2 | 11 | 9 | 15 | 1 | 10 | 3 | 4 | 8 | 14 | 3/6 |
| fir_il | 13 | 9 | 10 | 7 | 4 | 8 | 6 | 15 | 3 | 11 | 14 | 2 | 1 | 12 | 5 | 3/12 |
| fir_inf | 8 | 7 | 6 | 13 | 2 | 15 | 1 | 11 | 9 | 14 | 12 | 3 | 4 | 10 | 5 | 3/5 |
| fir_jelsr | 3 | 6 | 2 | 4 | 1 | 5 | 9 | 8 | 15 | 11 | 10 | 13 | 7 | 14 | 12 | 3/13 |
| fir_KW | 13 | 15 | 14 | 10 | 12 | 5 | 1 | 2 | 3 | 4 | 6 | 7 | 8 | 9 | 11 | 3/12 |
| fir_lapscore | 8 | 9 | 4 | 1 | 3 | 2 | 5 | 15 | 6 | 11 | 13 | 7 | 10 | 14 | 12 | 3/12 |
| fir_llc | 2 | 9 | 4 | 8 | 1 | 3 | 6 | 7 | 5 | 11 | 10 | 13 | 15 | 12 | 14 | 3/12 |
| fir_lnrd | 12 | 14 | 13 | 11 | 10 | 1 | 2 | 6 | 3 | 5 | 4 | 9 | 7 | 8 | 15 | 3/13 |
| fir_mat_ttest | 7 | 6 | 13 | 2 | 15 | 11 | 1 | 14 | 8 | 12 | 3 | 9 | 10 | 4 | 5 | 3/4 |
| fir_mat_entropy | 6 | 7 | 13 | 15 | 8 | 14 | 2 | 11 | 9 | 1 | 10 | 3 | 12 | 5 | 4 | 3/7 |
| fir_mat_bhy | 6 | 7 | 13 | 15 | 8 | 14 | 2 | 11 | 9 | 1 | 10 | 3 | 12 | 5 | 4 | 3/7 |
| fir_mat_roc | 6 | 7 | 5 | 13 | 12 | 11 | 15 | 2 | 1 | 8 | 14 | 9 | 3 | 10 | 4 | 3/8 |
| fir_mat_wilcoxon | 6 | 7 | 13 | 12 | 11 | 15 | 2 | 1 | 8 | 14 | 9 | 3 | 10 | 4 | 5 | 3/7 |
| fir_mat_relieff | 13 | 8 | 15 | 4 | 3 | 14 | 2 | 9 | 1 | 11 | 5 | 7 | 12 | 10 | 6 | 3/12 |
| fir_mat_lasso | 13 | 14 | 12 | 6 | 1 | 3 | 8 | 9 | 4 | 5 | 2 | 11 | 10 | 7 | 15 | 3/14 |
| fir_mc | 12 | 14 | 10 | 15 | 13 | 7 | 11 | 2 | 1 | 3 | 6 | 5 | 4 | 9 | 8 | 3/8 |
| fir_mmls | 9 | 13 | 5 | 15 | 8 | 7 | 10 | 2 | 4 | 11 | 14 | 1 | 12 | 3 | 6 | 3/8 |
| fir_nnsa | 10 | 11 | 1 | 2 | 14 | 6 | 5 | 4 | 9 | 8 | 7 | 15 | 13 | 3 | 12 | 3/14 |
| fir_ol | 3 | 1 | 2 | 6 | 8 | 4 | 9 | 5 | 7 | 10 | 11 | 14 | 12 | 13 | 15 | 3/14 |
| fir_pwfp | 2 | 1 | 3 | 4 | 5 | 8 | 6 | 7 | 9 | 10 | 11 | 12 | 13 | 14 | 15 | 3/13 |
| fir_ru | 14 | 15 | 10 | 13 | 7 | 12 | 11 | 1 | 6 | 2 | 5 | 3 | 4 | 9 | 8 | 3/10 |
| fir_sgo | 12 | 15 | 2 | 7 | 9 | 11 | 1 | 3 | 8 | 5 | 13 | 10 | 6 | 4 | 14 | 3/11 |
| fir_soc | 2 | 4 | 1 | 9 | 8 | 3 | 6 | 5 | 11 | 7 | 10 | 15 | 13 | 14 | 12 | 3/13 |

## IV. CONCLUSION

This study present a matlab toolbox for feature importance ranking and the toolbox is available online. A total of 30 FIR methods have been integrated that paves the way for feature selection and intelligent diagnosis in real-world applications. Further, an evaluation is conducted on an ultrasound image database for breast cancer diagnosis to show how to use the toolbox and to give clues on feature importance ranking and selection. At last, comparison of FIR methods in detail and how to measure the effective should be fully investigated in our future work.